\documentclass[final,5p,times,twocolumn]{elsarticle}

\usepackage[switch]{lineno}
\usepackage[hidelinks]{hyperref}
\modulolinenumbers[5]

\usepackage{amsmath}
\usepackage{amssymb}
\usepackage{paralist}
\usepackage{subcaption}
\usepackage{siunitx}
\usepackage{graphicx}
\usepackage{booktabs}
\usepackage{xcolor}
\usepackage{MnSymbol}
\usepackage{multirow}
\usepackage{threeparttable}
\usepackage{array}
\newcolumntype{C}[1]{>{\centering\let\newline\\\arraybackslash\hspace{0pt}}m{#1}}

\journal{Neurocomputing}

\begin{document}

\begin{frontmatter}

\title{MAMNet: Multi-path Adaptive Modulation Network for Image Super-Resolution}

\author{Jun-Hyuk Kim}
\ead{junhyuk.kim@yonsei.ac.kr}

\author{Jun-Ho Choi}
\ead{idearibosome@yonsei.ac.kr}

\author{Manri Cheon}
\ead{manri.cheon@yonsei.ac.kr}

\author{Jong-Seok Lee\corref{cor}}
\ead{jong-seok.lee@yonsei.ac.kr}

\address{School of Integrated Technology, Yonsei University, 85 Songdogwahak-ro, Yeonsu-gu, Incheon, Korea}

\cortext[cor]{Corresponding author}

\begin{abstract}
In recent years, single image super-resolution (SR) methods based on deep convolutional neural networks (CNNs) have made significant progress.
However, due to the non-adaptive nature of the convolution operation, they cannot adapt to various characteristics of images, which limits their representational capability and, consequently, results in unnecessarily large model sizes.
To address this issue, we propose a novel multi-path adaptive modulation network (MAMNet).
Specifically, we propose a multi-path adaptive modulation block (MAMB), which is a lightweight yet effective residual block that adaptively modulates residual feature responses by fully exploiting their information via three paths.
The three paths model three types of information suitable for SR:
\begin{inparaenum}[1)]
	\item channel-specific information (CSI) using global variance pooling, 
	\item inter-channel dependencies (ICD) based on the CSI,
	\item and channel-specific spatial dependencies (CSD) via depth-wise convolution.
\end{inparaenum}
We demonstrate that the proposed MAMB is effective and parameter-efficient for image SR than other feature modulation methods.
In addition, experimental results show that our MAMNet outperforms most of the state-of-the-art methods with a relatively small number of parameters.

\end{abstract}


\end{frontmatter}

\section{Introduction}
\label{section:intro}
Single image super-resolution (SR) is the process of inferring a high-resolution (HR) image from a single low-resolution (LR) image.
It is one of the computer vision problems progressing rapidly with the development of deep learning.
Recently, convolutional neural network (CNN)-based SR methods~\cite{dong2014learning, kim2016accurate,ledig2016photo, lim2017enhanced, tong2017image, haris2018deep, zhang2018residual, li2018multi, zhang2018rcan} have shown better performance compared with previous hand-crafted methods~\cite{chang2004neighbor, sun2008gradient, yang2010image, timofte2013anchored, huang2015single}.

Stacking an extensive amount of layers is a common practice to improve performance of deep networks~\cite{montufar2014number}.
After Kim et al.~\cite{kim2016accurate} first applying residual learning in their very deep CNN for SR (VDSR), this trend goes on for SR as well.
Ledig et al.~\cite{ledig2016photo} propose a deeper network (SRResNet) than VDSR based on the ResNet architecture.
Lim et al.~\cite{lim2017enhanced} modify SRResNet and propose two very large networks having superior performance: wider one and deeper one, i.e., enhanced deep ResNet for SR (EDSR) and multi-scale deep SR (MDSR), respectively.
In addition, there have been approaches adopting DenseNet~\cite{huang2017densely} for SR, e.g.,~\cite{tong2017image,zhang2018residual}.

While a huge size of CNN-based SR network tends to yield improved performance, it still has some limitations due to its non-adaptive nature, i.e., convolution
is performed with fixed weights regardless of the input.
First, most CNN-based methods internally treat all types of LR images equally, which may not effectively distinguish the detailed characteristics of the content (e.g., natural vs. computer-generated ones).
Second, all regions are considered equally within an LR image, which may not effectively distinguish the detailed characteristics of each region (e.g., low vs. high frequency).
These limitations restrict the representational capability of SR networks, which leads to inefficient parameter usage, i.e., excessively large model sizes.
Therefore, designing flexible networks for various situations is required for effective and efficient SR.

A few recent SR methods~\cite{zhang2018rcan, hu2018channel} attempt to address this issue.
They design adaptive SR networks by modulating convolutional feature responses utilizing their information.
Zhang et al.~\cite{zhang2018rcan} propose the residual channel attention block (RCAB) that modulates channel-wise feature responses by exploiting inter-channel dependencies. 
Hu et al.~\cite{hu2018channel} propose the channel-wise and spatial attention residual (CSAR) block to adaptively modulate feature responses by explicitly modelling channel-wise and spatial interdependencies.
However, these methods do not make full use of information from feature responses for imposing sufficient adaptability on SR networks.
Specifically, although the CSAR block exploits two types of interdependencies between feature responses, channel-specific information is not exploited for feature modulation. 

Furthermore, from a network design perspective, these methods do not fully take into account the characteristics of SR, which differ from those of high-level vision problems such as image classification.
First, both RCAB and the CSAR block use the squeeze-and-excitation (SE) block~\cite{hu2018squeeze} for modelling the inter-channel dependencies, which uses global average pooling to extract channel-wise statistics.
However, since image SR ultimately aims at restoring high-frequency components of images, it is more reasonable to exploit frequency-related information as statistics representing channels.
Second, the CSAR block models spatial interdependencies without considering channel-specific characteristics, i.e., the same spatial modulation is performed for all channels.
While such a feature modulation strategy is effective when certain spatial regions of an image are more important than others (for, e.g., image classification~\cite{woo2018cbam}), it is not suitable for image SR where all areas of images have similar importance.

\begin{figure*} [ht]
	\centering
	\includegraphics[width=\textwidth]{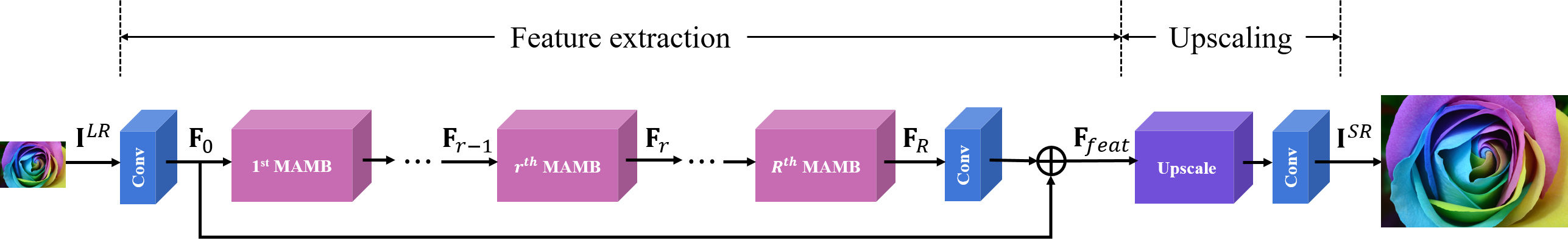}
	\caption{Overall architecture of our proposed network.} 
	\label{fig:architecture}
\end{figure*}

To address these issues, we propose a novel multi-path adaptive modulation network (MAMNet), whose overall architecture is illustrated in Figure~\ref{fig:architecture}. 
Specifically, we design a novel multi-path adaptive modulation block (MAMB) (Figure~\ref{fig:MAMB}), a lightweight yet effective residual block, which adaptively modulates residual feature responses by fully exploiting their information via three paths in a SR-optimized manner.
The three paths correspond to the three different types of information, i.e., channel-specific information (CSI), inter-channel dependencies (ICD), and channel-specific spatial dependencies (CSD).
For modelling CSI, we extract a statistic representing each channel by performing global variance pooling that can reflect frequency-related information, which is a more reasonable approach to SR compared to global average pooling.
To the best of our knowledge, this concept has not been adopted in existing image SR methods.
Based on the extracted channel-wise variances, we model ICD using two fully-connected layers.
Lastly, for modelling CSD, we generate a spatial modulation map for each channel via a depth-wise convolution layer.   
Compared to the previous methods~\cite{woo2018cbam,hu2018channel}, our method is effective for SR in that it models spatial dependencies with preserving the characteristics inherent to each channel.

In summary, our main contributions are as follows:
\begin{itemize}
	\item We propose a novel multi-path adaptive modulation network (MAMNet) for effective and parameter-efficient image SR.
	The proposed MAMNet resolves the non-adaptivity inherent in most of the previous CNN-based SR networks, by adaptively modulating convolutional feature responses.
	
	\item As the key component of our MAMNet, we propose a multi-path adaptive modulation block (MAMB) to fully exploit information of the feature responses for their modulation.
	The information exploitation proceeds via three paths corresponding to the three types of information, i.e., channel-specific information (CSI), inter-channel dependencies (ICD), and channel-specific spatial dependencies (CSD).
	
	\item We model the three types of information in a SR-optimized manner.
    First, we extract CSI by performing a global variance pooling that reflects frequency-related information. 
    In addition, we model CSD via a depth-wise convolution, which not only exploits spatial dependencies but also preserves channel-specific characteristics. 
\end{itemize}

The rest of this paper is organized as follows.
Section~\ref{section:related_works} reviews the deep CNN-based image SR methods and attention mechanisms in CNNs.
Section~\ref{section:proposed methods} introduces our proposed MAMNet in detail.
We discuss the differences between relevant studies and the proposed method in~Section~\ref{section:discussion}.
Section~\ref{section:Experiments} analyzes the proposed method in detail and provides performance comparisons with other methods experimentally.
Finally, we conclude our work in Section~\ref{section:conclusion}.

\section{Related Works}
\label{section:related_works}
Many CNN-based networks have been proposed to improve the performance of image SR~\cite{dong2014learning, kim2016accurate, ledig2016photo, lim2017enhanced, tong2017image, haris2018deep, zhang2018residual, li2018multi, zhang2018rcan}.
As mentioned in Section~\ref{section:intro}, they have evolved toward deepening networks.
We first review deep CNN-based SR networks developed in previous studies.

Our proposed method is related to the attention mechanism~\cite{hu2018squeeze}, which is one of the notable network structures to recalibrate the feature responses so that more adaptive and efficient
training is possible.
We briefly review the methods in which attention mechanisms are applied to CNNs.

\textbf{Deep CNN-based image SR.}
After Lim et al.~\cite{lim2017enhanced} proposing huge ResNet-based SR models, i.e., EDSR and MDSR, and Tong et al.~\cite{tong2017image} adopting the DenseNet structure for image SR, i.e., SRDenseNet, there have been some approaches to further improve the performance of image SR~\cite{haris2018deep,zhang2018residual,li2018multi,zhang2018rcan}.
Zhang et al.~\cite{zhang2018residual} propose a residual dense network (RDN) to fully exploit hierarchical features from LR images.
RDN consists of stacked residual dense blocks (RDBs), which combine the ResNet and DenseNet structures.
Haris et al.~\cite{haris2018deep} propose deep back-projection networks (DBPNs), which exploit the mutual dependencies of LR and HR images.
Inspired by~\cite{irani1991improving}, while most recent models use the post up-sampling approach, DBPNs consist of iterative up- and down-sampling layers to explicitly model an error feedback mechanism.   
Inspired by the inception block~\cite{szegedy2015going}, Li et al.~\cite{li2018multi} propose the multi-scale residual network (MSRN), which employs different convolution kernels with different sizes in its basic building block.
The aforementioned deep CNN-based SR methods have achieved good performance through various network structures.
However, they treat all types of information equally and are not adaptable to various situations.

\textbf{Attention mechanism.}
In the cognitive process of human, the use of selective information, i.e., focusing on more important information, generally occurs~\cite{mnih2014recurrent}.
This process is referred to as the attention mechanism, which is widely used in various applications~\cite{xu2015show,yao2015describing}.
Recently, there are some approaches to apply attention mechanisms to ResNet-based networks~\cite{wang2017residual,hu2018squeeze,woo2018cbam,zhang2018rcan,hu2018channel}.
Wang et al.~\cite{wang2017residual} propose a residual attention network (RAN) to improve the performance of image classification.
Since their attention module generates 3D attention maps using 3D feature maps directly, it is helpful to improve performance, but is relatively heavy.
Hu et al.~\cite{hu2018squeeze} introduce a compact attention mechanism, i.e., the SE block, which adaptively recalibrates 3D feature maps by explicitly modelling inter-channel dependencies.
The SE block generates 1D channel attention maps using only 1D global average pooled features.
While the attention mechanism in the SE block uses only inter-channel relation for refining feature maps, the convolutional block attention module (CBAM)~\cite{woo2018cbam} exploits both inter-channel and spatial relationships of feature maps through its channel and spatial attention modules, respectively, which are performed sequentially.
The channel attention module in CBAM is different from the SE block in that global max pooling is additionally performed to extract multiple channel statistics.

As mentioned in Section~\ref{section:intro}, while Zhang et al.~\cite{zhang2018rcan} and Hu et al.~\cite{hu2018channel} try to apply the attention mechanism to image SR, they do not fully exploit feature responses and different characteristics between high- and low-level computer vision problems are not adequately considered.
Differences between these methods and ours are explained in Section~\ref{section:discussion}.

\section{Proposed Method}
\label{section:proposed methods}
\subsection{Network Architecture}
The overall architecture of our MAMNet is illustrated in Figure~\ref{fig:architecture}.
It can be divided into two parts: 
\begin{inparaenum}[1)]
	\item feature extraction part, and
	\item upscaling part.
\end{inparaenum}
Let $\textbf{I}^{LR}\in\mathbb{R}^{H\times W\times3}$ and $\textbf{I}^{SR}$ denote the input LR image and the corresponding output image, respectively.
At the beginning, one convolution layer is applied to $\textbf{I}^{LR}$ to extract initial feature maps, i.e.,
\begin{equation}
\textbf{F}_{0} = f_{0}(\textbf{I}^{LR}),
\end{equation}
where $f_{0}(\cdot)$ denotes the first convolution and $\textbf{F}_{0}$ means the extracted feature maps to be fed into the first MAMB, which is described in detail in Section~\ref{section:MAMB}.
$\textbf{F}_{0}$ is updated through $R$ MAMBs and one convolution layer.
Then, the updated feature maps are added to $\textbf{F}_{0}$ by using the global skip connection:
\begin{equation}
\textbf{F}_{feat} = \textbf{F}_{0} + f_{feat}(\textbf{F}_{R}),
\end{equation}
where $\textbf{F}_{R}$ is the output feature maps of the $R$-th MAMB and $f_{feat}(\cdot)$ and $\textbf{F}_{feat}$ are the last convolution layer and feature maps of the feature extraction part, respectively.

For the upscaling part, we use the sub-pixel convolution layers~\cite{shi2016real}, which are followed by one convolution layer for reconstruction:
\begin{equation}
\textbf{I}^{SR} = f_{recon}(f_{up}(\textbf{F}_{feat})),
\end{equation}
where $f_{up}(\cdot)$ and $f_{recon}(\cdot)$ are the functions for upscaling and reconstruction, respectively.   

\begin{figure}[t]
	\centering
	\includegraphics[width=0.5\textwidth]{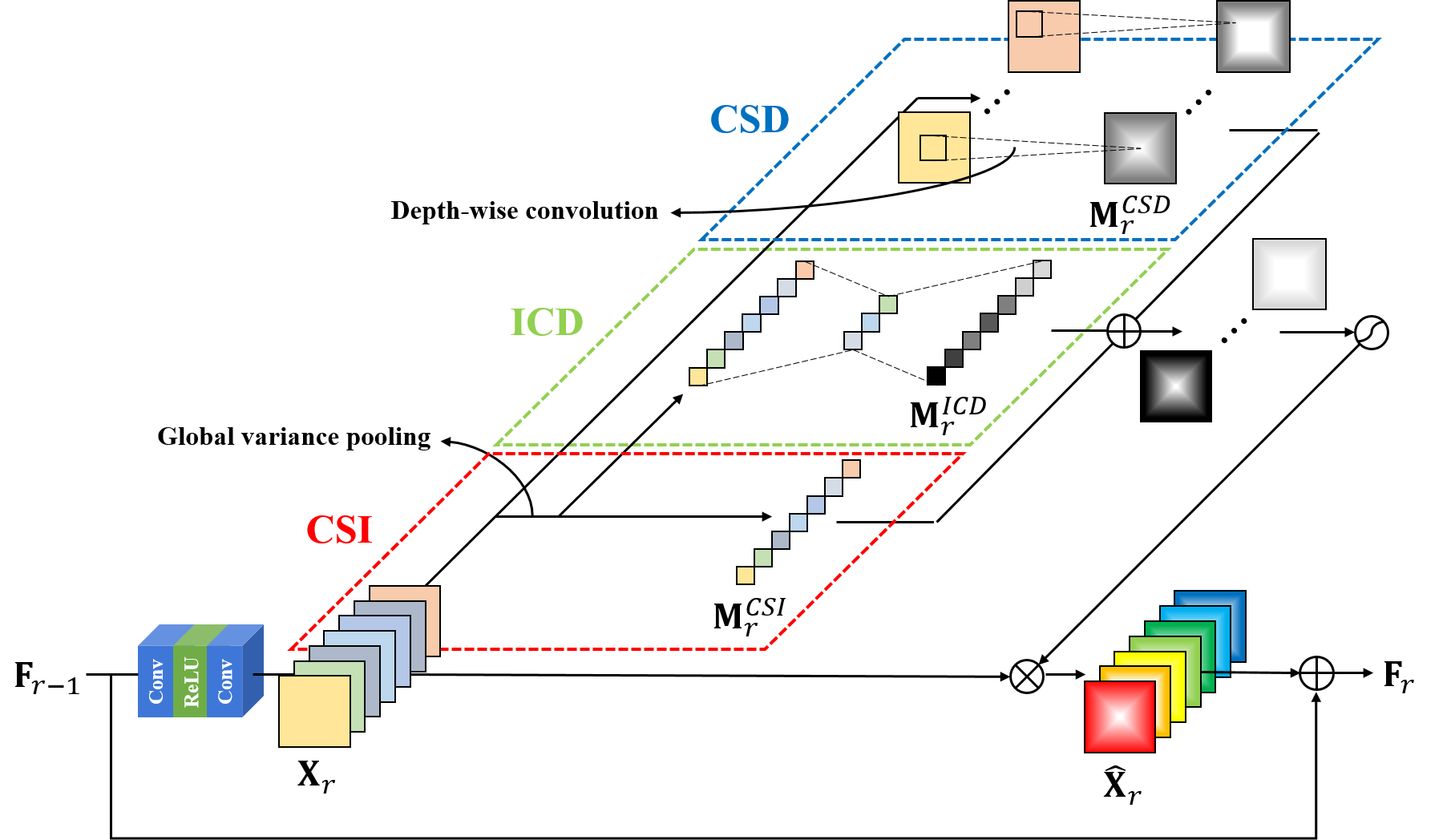}
	\caption{Multi-path adaptive modulation block (MAMB).
	} 
	\label{fig:MAMB}
\end{figure}

\subsection{Multi-path Adaptive Modulation Block}
\label{section:MAMB}
The structure of MAMB is illustrated in Figure~\ref{fig:MAMB}.
Let $\textbf{F}_{r-1}$ and $\textbf{F}_{r}$ be the input and output feature maps of the $r$-th MAMB.
Then, $\textbf{F}_{r}$ can be formulated as 
\begin{equation}
\begin{aligned}
\textbf{F}_{r} = f_{MAMB,r}(\textbf{F}_{r-1}) = \textbf{F}_{r-1} + f_{MAM}(\textbf{X}_{r}),
\label{eq:MAMNet}
\end{aligned}
\end{equation}
where $f_{MAMB,r}(\cdot)$ denotes the operations of the $r$-th MAMB, $\textbf{X}_{r}$ is the resultant feature maps having spatial dimensions $H\times W$ and a channel dimension $C$ after sequentially applying convolution, ReLU, and convolution on $\textbf{F}_{r-1}$, and $f_{MAM}(\cdot)$ means our multi-path adaptive modulation (MAM) that simultaneously exploits the three types of information, i.e., CSI, ICD, and CSD.
The feature modulation is performed as follows:
\begin{equation}
\begin{aligned}
\hat{\textbf{X}}_{r} &= f_{MAM}(\textbf{X}_{r}) \\
&= \sigma(\textbf{M}^{CSI}_{r}\oplus \textbf{M}^{ICD}_{r}\oplus \textbf{M}^{CSD}_{r}) \otimes \textbf{X}_{r},
\end{aligned}
\end{equation}
where $\textbf{M}^{CSI}_{r}$, $\textbf{M}^{ICD}_{r}$, and $\textbf{M}^{CSD}_{r}$ are the modulation maps using CSI, ICD, and CSD, respectively, $\oplus$ and $\otimes$ denote element-wise addition and multiplication, respectively, $\sigma(\cdot)$ is the sigmoid activation function,
and $\hat{\textbf{X}}_{r}$ is the modulated feature maps.
In order to enable the element-wise addition, $\textbf{M}^{CSI}_{r}$ and $\textbf{M}^{ICD}_{r}$ are resized to the size of $\textbf{M}^{CSD}_{r}$ via nearest-neighbor interpolation.

\textbf{Channel-specific information (CSI).}
Each channel of $\textbf{X}_{r}$ is the responses to a particular filter, which tend to vary depending on the characteristics of LR images.
Therefore, we utilize the CSI to adaptively modulate each channel.
It is important to extract a statistic that can effectively represent the characteristics of each channel.
Since image SR ultimately aims at restoring high-frequency components of images, we choose to use the variance, a frequency-related measure, for modelling the CSI.
Given $\textbf{X}_{r}=[\textbf{x}_{r,1},\textbf{x}_{r,2},...,\textbf{x}_{r,C}]$, the $c$-th modulation map $\textbf{m}^{CSI}_{r,c}$ of $\textbf{M}^{CSI}_{r}$ is calculated by:
\begin{equation}
\mu^{CSI}_{r,c}=\frac{1}{H\times W}\sum_{i=1}^{H}\sum_{j=1}^{W} \textbf{x}_{r,c}(i,j),
\end{equation}
\begin{equation}
\textbf{m}^{CSI}_{r,c}=\frac{1}{H\times W}\sum_{i=1}^{H}\sum_{j=1}^{W} (\textbf{x}_{r,c}(i,j)-\mu^{CSI}_{r,c})^{2},
\end{equation}
where the modulation map is used after standardization.

\textbf{Inter-channel dependencies (ICD).}
An LR image shows different interdependencies between channels depending on the types of textures it contains~\cite{gatys2015texture}.
For example, an image with a repeated pattern shows high interdependencies among channels related to the pattern.
For generating the modulation map $\textbf{M}_{r}^{ICD}$, we exploit this information, i.e., ICD, by employing two fully-connected layers whose structure is the same to that in~\cite{hu2018squeeze, hu2018channel}: 
\begin{equation}
\textbf{M}_{r}^{ICD}=\textbf{W}_{2}\delta(\textbf{W}_{1}\textbf{M}_{r}^{CSI}),
\label{eq:excitation}
\end{equation}
where $\textbf{W}_{1}\in\mathbb{R}^{\frac{C}{16}\times~C}$ and $\textbf{W}_{2}\in\mathbb{R}^{C\times \frac{C}{16}}$ are the parameters of the fully-connected layers, and $\delta(\cdot)$ is the ReLU activation function.

\textbf{Channel-specific spatial dependencies (CSD).}
Each channel in the feature maps $\textbf{X}_{r}$ has a different meaning depending on the role of the filter used.
For example, some filters may extract the edge components in the horizontal direction, and some other filters may extract the edge components in the vertical direction.
From the viewpoint of SR, where it is important to extract as much information as possible from LR images, it is expected that every channel plays its own important role.
In addition, the importance of the channels varies spatially.
For example, in the case of edges or complex textures, detailed information, i.e., those from complex filters, would be important.
On the other hand, in the case of the region having almost no high-frequency components such as sky or homogeneous areas of comic images, relatively less detailed information would be more important and need to be attended.
Therefore, it is necessary to model spatial interdependencies within each channel, i.e., CSD.
To preserve channel-specific characteristics, we obtain a different 2D modulation map for each channel using depth-wise convolution~\cite{howard2017mobilenets}, where independent convolution operations are performed for each channel:
\begin{equation}
\textbf{M}^{CSD}_{r} = f_{depth}(\textbf{X}_{r}),
\end{equation}
where $f_{depth}(\cdot)$ denotes the $3\times 3$ depth-wise convolution.




\section{Discussion}
\label{section:discussion}
\textbf{Difference from the SE block and RCAB.}
The residual channel attention network (RCAN)~\cite{zhang2018rcan} adopts the channel attention mechanism in RCAB, which is the same to the SE block~\cite{hu2018squeeze}. 
It relies on global average pooling to extract representative statistics. 
However, it is designed for high-level computer vision tasks such as image classification, and thus may not be optimal for image SR. 
We propose a variance-based channel modulation using ICD to improve the SR performance, as will be shown experimentally (Table~\ref{tab:ablation_icd}). 
In addition, we exploit not only ICD but also CSI and CSD, which leads to fully utilize the information of residual feature maps.

\textbf{Difference from CBAM.}
The channel attention module in CBAM~\cite{woo2018cbam} uses both global average and max pooling, achieving performance improvement in high-level computer vision tasks. 
However, as will be shown later (Table~\ref{tab:ablation_icd}), the additional use of max pooling lowers SR performance, indicating that it is important to choose appropriate statistics according to the application.
We demonstrate that the frequency-related statistic, i.e., variance, is effective for image SR.
In addition, it should be noted that the spatial attention module in CBAM and our method modelling CSD are largely different.
After compressing the information via global average and max pooling in the channel direction, CBAM generates a single 2D spatial attention map through a convolution layer.
This approach has two drawbacks in SR:
Each channel has different information (e.g., frequency-related information) and plays a specific role. 
Therefore, it is not reasonable to squeeze information through pooling. 
In addition, it is not suitable for SR to apply a single 2D spatial attention map without reflecting channel-specific characteristics. 
As shown in Table~\ref{tab:ablation_csd}, this method causes performance degradation.
On the other hand, our method is applied for each channel to modulate it in a spatially adaptive manner.
Furthermore, CBAM does not fully exploit information of feature responses, because it does not consider CSI for the feature modulation.

\textbf{Difference from the CSAR block.}
The CSAR block~\cite{hu2018channel} also employs the SE block for modelling ICD.
In addition, it generates a single 2D map like CBAM for modelling spatial interdependencies, while in our MAMB, we model CSD instead.
As shown in Table~\ref{tab:ablation_csd}, this approach degrades the SR performances.
It should be noted that such results are obtained with employing overly large numbers of parameters because it does not adequately consider the characteristics of image SR.
Moreover, the CSAR block also does not utilize CSI for feature modulation.  

\section{Experiments}
\label{section:Experiments}
\textbf{Datasets and metrics.}
In our experiments, we follow the evaluation protocol commonly used in many previous studies~\cite{lim2017enhanced, zhang2018residual,zhang2018rcan}.
We train all our models using the training images from the DIV2K dataset~\cite{Timofte_2018_CVPR_Workshops}.
It contains 800 RGB HR training images and their corresponding LR training images.
For evaluation, we use five datasets commonly used in SR benchmarks: Set5~\cite{bevilacqua2012low}, Set14~\cite{zeyde2010single}, BSD100~\cite{martin2001database}, Urban100~\cite{huang2015single}, Manga109~\cite{matsui2017sketch}.
The Set5, Set14, and BSD100 datasets consist of natural images.
The Urban100 dataset includes images related to building structures with complex and repetitive patterns, which are challenging for SR.
The Manga109 dataset consists of images taken from Japanese cartoon, which are computer-generated images and have different characteristics from natural ones.
To evaluate SR performance, we calculate the peak signal-to-noise ratio (PSNR) and structural similarity (SSIM) index on the Y channel after converting to YCbCr channels.

\textbf{Implementation Details.}
To construct an input mini-batch for training, we randomly crop a 48$\times$48 patch from each of the randomly selected 16 LR training images.
For data augmentation, the patches are randomly horizontal-flipped and rotated (\ang{90}, \ang{180}, and \ang{270}).
Before feeding the mini-batch into our networks, we subtract the average value of the entire training images for each RGB channel of the patches.
We set the size and number of filters as 3$\times3$ and 64 respectively in all convolution layers except those for the upscaling part.
All our networks are optimized using the Adam optimizer~\cite{kingma2014adam} to minimize the L1 loss function, where the parameters of the optimizer are set as $\beta1=0.9$, $\beta2=0.999$, and $\epsilon=10^{-8}$.
The learning rate is initially set to $10^{-4}$, which decreases by a half at every $2\times10^{5}$ iterations.  
We implement our networks using the Tensorflow framework with NVIDIA GeForce GTX 1080 GPU\footnote{Our code is publicly available at \url{https://github.com/junhyukk/MAMNet-tensorflow}}.



\begin{table}[t]
	\begin{center}
		\scalebox{0.7}{
			\begin{tabular}{lccccc}
				\toprule
				Methods & Set5 & Set14 & BSD100 & Urban100 & Manga109\\	
				\midrule
				Baseline & 37.90 & 33.58 & \textcolor{red}{32.17} & 32.13 & 38.47\\
				+ Max & 37.94 & 33.53 & 32.16 & 32.09 & 38.44\\
				+ Avg & \textcolor{red}{37.96} & 33.59 & 32.07 & 32.24 & \textcolor{blue}{38.65}\\
				+ Var & 37.93 & \textcolor{blue}{33.61} & \textcolor{red}{32.17} & \textcolor{blue}{32.28} & 38.64\\
				+ Power & 37.92 & 33.59 & 32.16 & 32.20 & 38.51\\
				+ Standardized var (Ours) & \textcolor{blue}{37.95} & \textcolor{red}{33.63} & \textcolor{red}{32.17} & \textcolor{red}{32.33} & \textcolor{red}{38.73}\\
                \bottomrule	
			\end{tabular}
		}
	\end{center}
	\caption{Effect of using different pooling methods for CSI. Average PSNR values (dB) for $\times$2 SR on the five datasets are shown. 
	Red and blue colors indicate the best and second best performance for each dataset, respectively.}
	\label{tab:ablation_csi}
\end{table}

\begin{table}[t]
	\begin{center}
		\scalebox{0.65}{
			\begin{tabular}{lccccc}
				\toprule
				Methods & Set5 & Set14 & BSD100 & Urban100 & Manga109\\	
				\midrule
				Baseline & 37.90 & \textcolor{blue}{33.58} & \textcolor{red}{32.17} & 32.13 & 38.47\\
				+ CAM of CBAM~\cite{woo2018cbam} (Max \& Avg) & \textcolor{black}{37.91} & \textcolor{black}{33.51} & \textcolor{black}{32.14} & 32.14 & \textcolor{black}{38.19}\\			
				+ RCAB~\cite{zhang2018rcan} (Avg) & \textcolor{blue}{37.96} & \textcolor{blue}{33.58} & \textcolor{red}{32.17} & 32.24 & \textcolor{black}{38.60}\\
				+ Var & 37.93 & \textcolor{black}{33.55} & \textcolor{red}{32.17} & \textcolor{blue}{32.26} & \textcolor{blue}{38.67}\\
				+ Standardized var (Ours) & \textcolor{red}{37.97} & \textcolor{red}{33.66} & \textcolor{red}{32.17} & \textcolor{red}{32.32} & \textcolor{red}{38.71}\\
                \bottomrule	
			\end{tabular}
		}
	\end{center}
	\caption{Effect of using different CSI for modelling ICD. Average PSNR values (dB) for $\times$2 SR on the five datasets are shown. 
	Red and blue colors indicate the best and second best performance for each dataset, respectively.}
	\label{tab:ablation_icd}
\end{table}

\begin{table}[t]
	\begin{center}
		\scalebox{0.63}{
			\begin{tabular}{lcccccc}
				\toprule
				Methods & \# params. (K) & Set5 & Set14 & BSD100 & Urban100 & Manga109\\	
				\midrule
				Baseline & 1370 & 37.90 & \textcolor{blue}{33.58} & \textcolor{red}{32.17} & \textcolor{red}{32.13} & \textcolor{red}{38.47}\\
				SAM of CBAM~\cite{woo2018cbam} & 1371 & 37.84 & 33.52 & 32.12 & 31.93 & 38.31\\
				SA of CSAR block~\cite{hu2018channel} & 1505 & \textcolor{blue}{37.91} & 33.56 & 32.14 & 32.02 & 38.33\\
				Ours (CSD) & 1380 & \textcolor{red}{37.95} & \textcolor{red}{33.59} & \textcolor{red}{32.17} & \textcolor{red}{32.13} & \textcolor{blue}{38.46}\\
                \bottomrule	
			\end{tabular}
		}
	\end{center}
	\caption{Performance and parameter efficiency of modelling CSD. Average PSNR values (dB) for $\times$2 SR on the five datasets are shown. 
	Red and blue colors indicate the best and second best performance for each dataset, respectively.
	}
	\label{tab:ablation_csd}
\end{table}

\begin{table}[t]
	\begin{center}
		\scalebox{0.7}{
			\begin{tabular}{lcccccccc}
				\toprule
				Components & \multicolumn{8}{c}{Different combinations of CSI, ICD, and CSD}\\	
				\midrule
				CSI & $\times$ & $\checkmark$ & $\times$ & $\times$ & $\checkmark$ & $\checkmark$ & $\times$ & $\checkmark$\\
				ICD & $\times$ & $\times$ & $\checkmark$ & $\times$ & $\checkmark$ & $\times$ & $\checkmark$ & $\checkmark$\\
				CSD & $\times$ & $\times$ & $\times$ & $\checkmark$ & $\times$ & $\checkmark$ & $\checkmark$ & $\checkmark$\\
				\midrule
				Set5 & 37.90 & 37.95 & 37.97 & 37.95 & 37.97 & 37.97 & \textcolor{blue}{37.98} & \textcolor{red}{37.99}\\
				Set14 & 33.58 & 33.63 & \textcolor{blue}{33.66} & 33.59 & \textcolor{red}{33.71} & 33.63 & 33.64 & \textcolor{black}{33.64}\\
				BSD100 & 32.17 & 32.17 & 32.17 & 32.17 & 32.17 & \textcolor{blue}{32.18} & \textcolor{blue}{32.18} & \textcolor{red}{32.19}\\
				Urban100 & 32.13 & 32.33 & 32.32 & 32.13 & 32.34 & \textcolor{blue}{32.35} & 32.34 & \textcolor{red}{32.38}\\
				Manga109 & 38.47 & 38.73 & 38.71 & 38.46 & \textcolor{blue}{38.75} & 38.72 & \textcolor{blue}{38.75} & \textcolor{red}{38.80}\\
				Average & 34.40 & 34.55 & 34.54 & 34.40 & \textcolor{blue}{34.56} & \textcolor{blue}{34.56} & \textcolor{blue}{34.56} & \textcolor{red}{34.59}\\
				\midrule
				$\#$ params. (K) & 1370 & 1370 & 1379 & 1380 & 1379 & 1380 & 1389 & 1389\\	
				$\#$ params. $\uparrow$ ($\%$) & - & 0 & 0.68 & 0.75 & 0.68 & 0.75 & 1.43 & 1.43\\				
				\bottomrule	
			\end{tabular}
		}
	\end{center}
	\caption{Ablation study on effects of CSI, ICD, and CSD. Average PSNR values (dB) for $\times$2 SR on the five datasets are shown. The row ``Average'' means the average PNSR values for all images in the five datasets. Red and blue colors indicate the best and second best performance for each dataset, respectively. The row ``$\#$ params. $\uparrow$'' shows the ratio of the increased number of parameters.}
	\label{tab:ablation}
\end{table}

\begin{figure*}[t]
	\centering
	\begin{subfigure}[b]{\textwidth}
		\includegraphics[width=\textwidth]{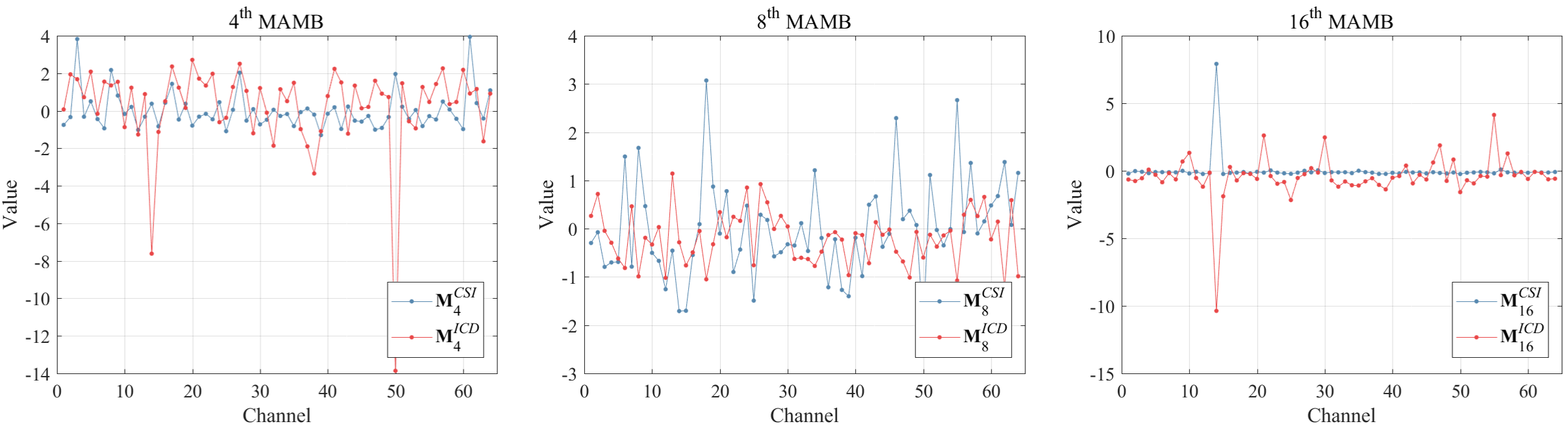}
		\caption{$\textbf{M}^{CSI}$ and $\textbf{M}^{ICD}$.}
		\label{fig:CIS_ICD}
	\end{subfigure}
	\begin{subfigure}[b]{\textwidth}
		\includegraphics[width=\textwidth]{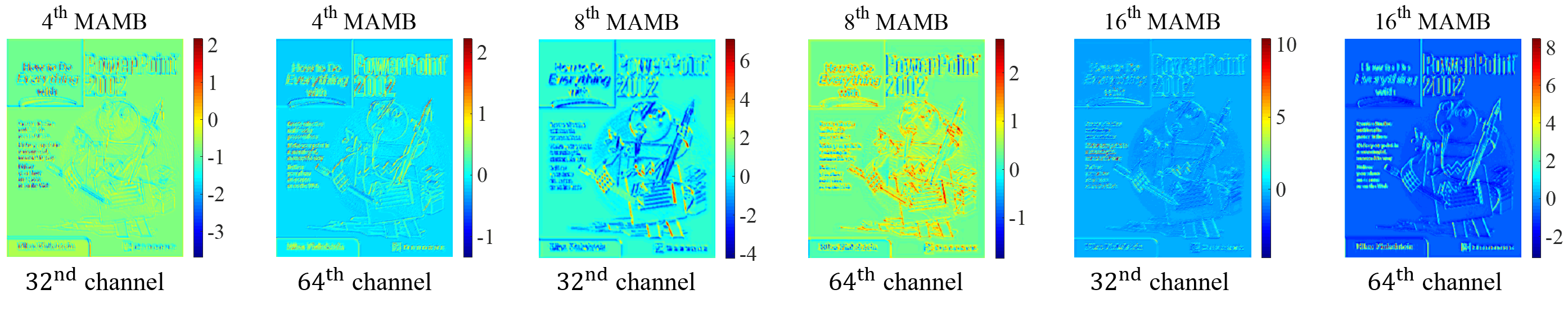}
		\caption{$\textbf{M}^{CSD}$.}
		\label{fig:CSD}
	\end{subfigure}
	\caption{Visualization of each path of our proposed multi-path adaptive modulation for ``ppt3'' from Set14~\cite{zeyde2010single}.}
	\label{fig:analysis_multi_path}
\end{figure*}

\subsection{Model Analysis}
\label{section:analysis}
In this section, we analyze the three paths (i.e., CSI, ICD, and CSD) of our proposed MAMB.
In MAMB, employing one or multiple path(s), except for CSI, increases the number of model parameters, and as the network becomes deeper, the number of such additional parameters becomes large.
We want to minimize the possibility to obtain improved performance simply due to such an increased number of parameters and, as a result, analyze the effect of our methods fairly. 
To this end, we conduct experiments with networks having 64 filters ($C$=64) and 16 residual blocks ($R$=16), which are not too deep or wide.
In addition, we analyze convergence of MAMNet with various configurations of $R$ and $C$.

\textbf{CSI.}
We examine the effectiveness of modelling CSI for image SR.
For the experiment, we construct six networks, one without feature modulation (i.e., baseline) and the rest with only a CSI path using different global pooling methods for modelling CSI, which are the maximum, average, variance, standardized variance, and power.
Here, ``power'' means the average of squared channel responses. 
Table~\ref{tab:ablation_csi} shows the result.
Unlike the result in the image classification task~\cite{woo2018cbam}, the max pooling rather degrades SR performance, which shows the importance of using appropriate methods for modelling CSI depending on the nature of the problem being solved.
For all datasets, our standardized variance-based method shows the best or second best SR performances, which demonstrates that using the frequency-related measure for CSI is effective for image SR. 

\textbf{ICD.}
We analyze whether the proposed variance-based CSI is also effective in modelling ICD for image SR.
We compare our method with the channel attention module (CAM) of CBAM and RCAB.
To model ICD, all the methods employ two fully-connected layers having the same structure.
The difference between them is that which CSI is used for modelling ICD. 
Similar to the result of Table~\ref{tab:ablation_csi}, the proposed method shows the best SR performance for all datasets in Table~\ref{tab:ablation_icd}, which strengthens the above conclusion that it is important to represent CSI appropriately.
In addition, using both average and max pooling shows lower SR performance compared to the baseline, which means that the max pooling is not helpful for image SR. 

\textbf{CSD.}
We compare the proposed CSD with the previous methods, the spatial attention module (SAM) of CBAM and the spatial attention (SA) unit of the CSAR block, both of which model spatial feature interdependencies without considering channel-specific characteristics.
Table~\ref{tab:ablation_csd} shows the result.
CBAM shows lower SR performance than the baseline for all datasets.
The CSAR block shows better SR performance than CBAM, but it is still worse than the baseline for all datasets except for Set5.
On the other hand, our method maintains similar or better performance in comparison to the baseline, which demonstrates that it is more effective to consider channel-specific characteristics when modelling spatial interdependencies.
Note that our method uses far fewer additional parameters than the CSAR block (0.7\% vs. 9.9\%).

\textbf{MAMB.}
Table~\ref{tab:ablation} shows the ablation study on the three paths (CSI, ICD, and CSD) of the proposed MAMB.
First, without CSI, ICD and CSD, the network exhibits the worst performance on average for the five datasets (34.40 dB), which implies the non-adaptive SR network does not effectively extract features from LR images.
This demonstrates that simply stacking residual blocks leads to limited representational power of deep networks for image SR.


Then, we add one of the three paths to the baseline (the second, third, and fourth columns of Table~\ref{tab:ablation}).
We confirm that CSI and ICD effectively lead to performance improvement (+0.15 dB and +0.14 dB, respectively) on average with no and negligible increase of the number of parameters (0$\%$ and 0.68$\%$, respectively).
CSD also yields the SR performances similar to or slightly better (+0.05 dB for Set5) than the baseline. 

In addition, we experiment on three networks using two of the three paths (the fifth, sixth, and seventh columns of Table~\ref{tab:ablation}).
We observe that the networks perform better than those using only one path.
Then, the best performance is achieved when all the three paths are used simultaneously, which is shown in the last column of Table~\ref{tab:ablation}.
The experimental results demonstrate that feature modulation exploiting sufficient information (CSI, ICD, and CSD) via multiple paths is effective for image SR.
Moreover, the performance improvement is achieved in a parameter-efficient manner (+0.19 dB with only 1.43$\%$ additional parameters).

To further analyze the role of each path of the proposed MAMB, we visualize the modulation map of each path in Figure~\ref{fig:analysis_multi_path}.
Figure~\ref{fig:analysis_multi_path}a shows those corresponding to CSI and ICD in the fourth, eighth, and last MAMBs, respectively.
Here, we have two interesting observations.
First, when the values of $\textbf{M}^{CSI}$ are similar across the channels, which means that the channels contain similar amounts of information, the values of $\textbf{M}^{ICD}$ vary significantly from channel to channel (the left and right panels of Figure~\ref{fig:analysis_multi_path}a).
Second, when $\textbf{M}^{ICD}$ is hardly activated differently across the channels, the values via $\textbf{M}^{CSI}$ are largely different depending on the channel and the information of the CSI is used dominantly for feature modulation, as shown in the middle panel of Figure~\ref{fig:analysis_multi_path}a.
These observations imply that although both CSI and ICD are derived from the channel-wise pooling, they have their own roles, which appear even complementary, for adaptive feature modulation.
Figure~\ref{fig:analysis_multi_path}b shows the 32nd and 64th channels of $\textbf{M}^{CSD}$ in each MAMB of Figure~\ref{fig:analysis_multi_path}a.
Each map has spatially varying values, which demonstrates that each map spatially modulates each channel adaptively.
In addition, the distribution of the map is different for each channel.
For example, the 64th channel of the 4th MAMB has a modulation map with relatively similar values in the spatial domain, which means that CSI and ICD can provide sufficient information for feature modulation.
On the other hand, the 32nd channel of the 16th MAMB has a modulation map with largely different values depending on the spatial location.
Specifically, the left area containing text has relatively large values, while the rest shows small values.
Through these results, it is confirmed that each channel requires different spatial modulation depending on its characteristics, i.e., CSD.

\begin{figure}[t]
	\centering
	\includegraphics[width=0.45\textwidth]{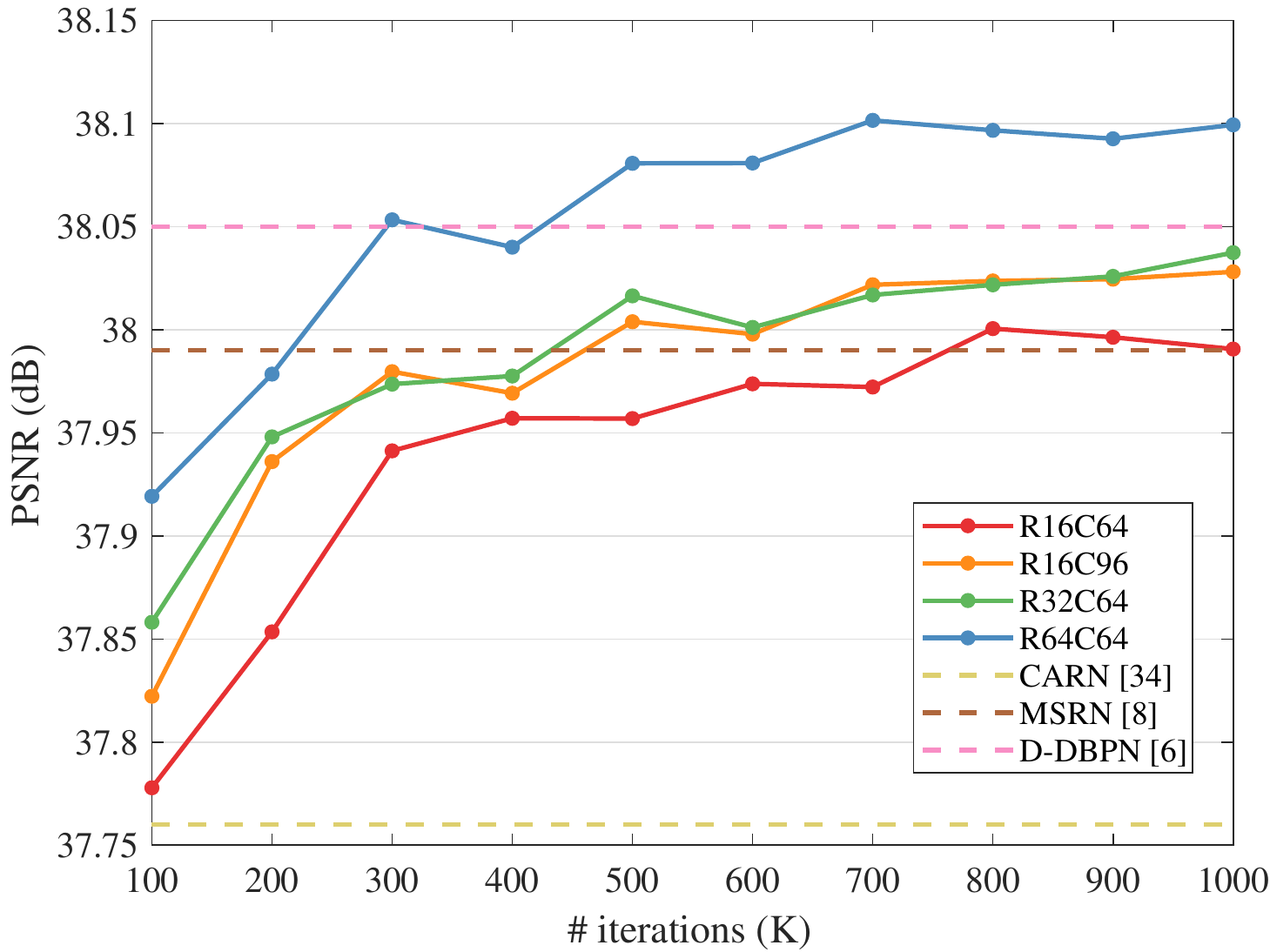}
	\caption{Convergence analysis of our models with various configurations for 
	$\times$2 SR on Set5~\cite{bevilacqua2012low}.}
	\label{fig:study}
\end{figure}

\textbf{Effect of $R$ and $C$.}
The structure of our MAMNet is determined by the number of MAMB ($R$) and the number of channels ($C$) used in each MAMB.
In this experiment, we examine the effect of these two variables on performance.
Starting from the case with $R=16$ and $C=64$ ($R16C64$), we increase $R$ or $C$.
The convergence analysis of the networks with different configurations according to the number of training iterations is shown in Figure~\ref{fig:study}, where CARN~\cite{ahn2018fast}, MSRN~\cite{li2018multi}, and D-DBPN~\cite{haris2018deep} are compared as references.
A larger value of $R$ or $C$ leads to performance improvement, which means that our proposed method allows deeper and wider structures through effective feature modulation.

\begin{figure}[t]
	\centering
	\includegraphics[width=0.45\textwidth]{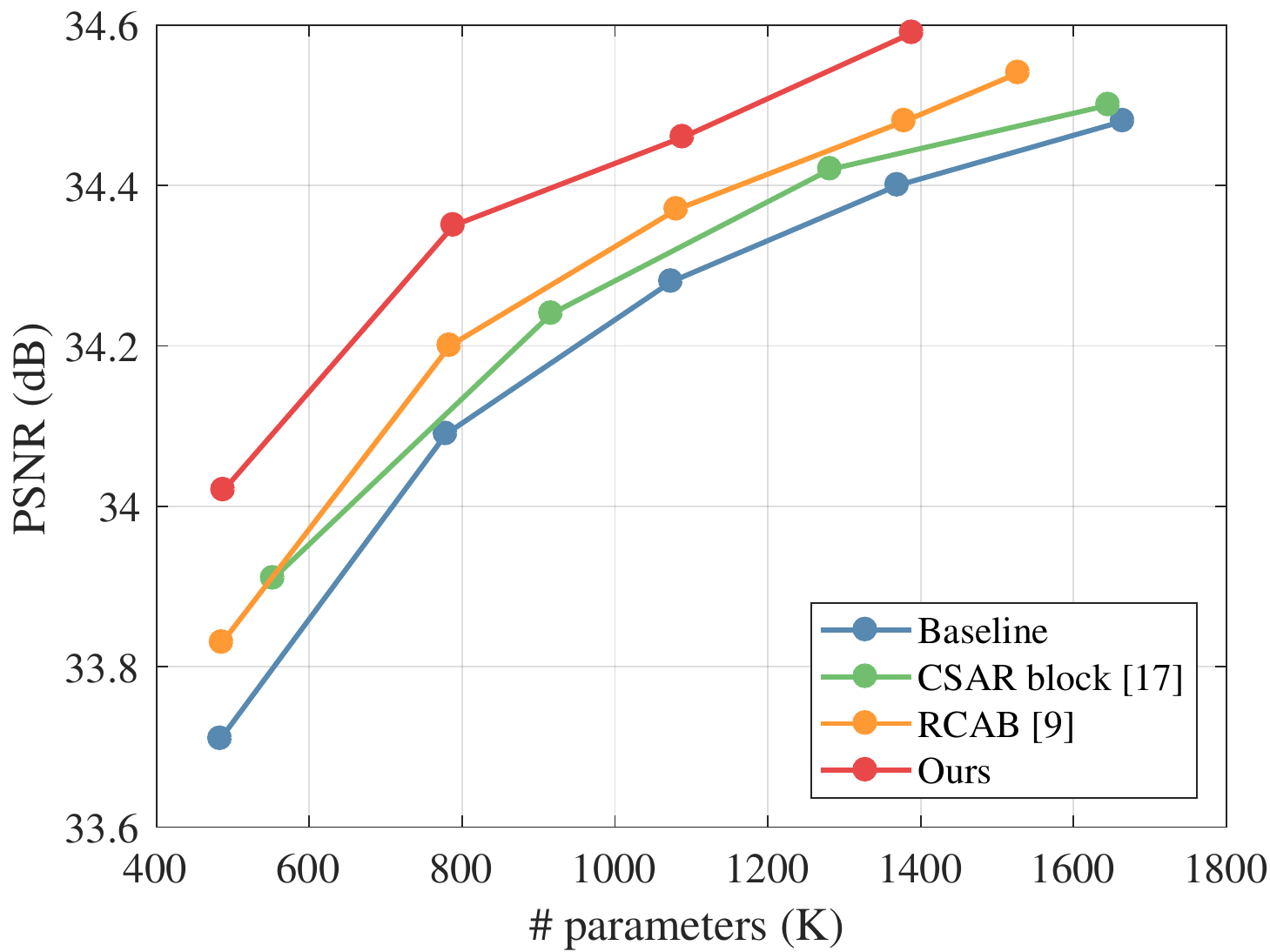}
	\caption{Comparison of differnet feature modulation methods for $\times$2 SR. 
	The PSNR values are the average values for all images of the five datasets.}
	\label{fig:attention_comp}
\end{figure}

\begin{figure*}[t]
	\centering
	\begin{subfigure}[b]{0.9\textwidth}
		\includegraphics[width=\textwidth]{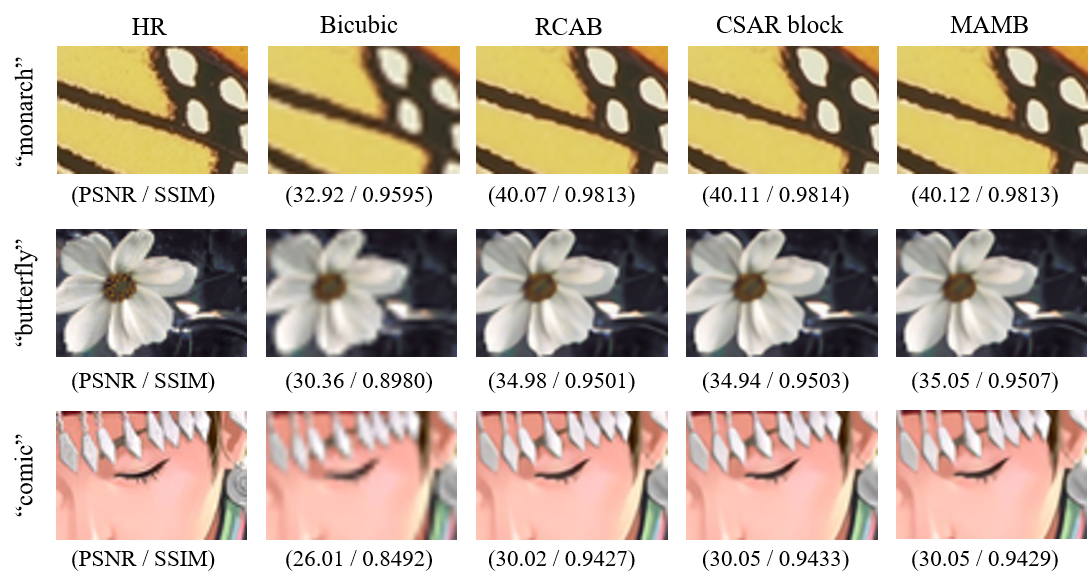}
		\caption{Easy cases.}
		\label{fig:easy}
	\end{subfigure}
	\begin{subfigure}[b]{0.9\textwidth}
		\includegraphics[width=\textwidth]{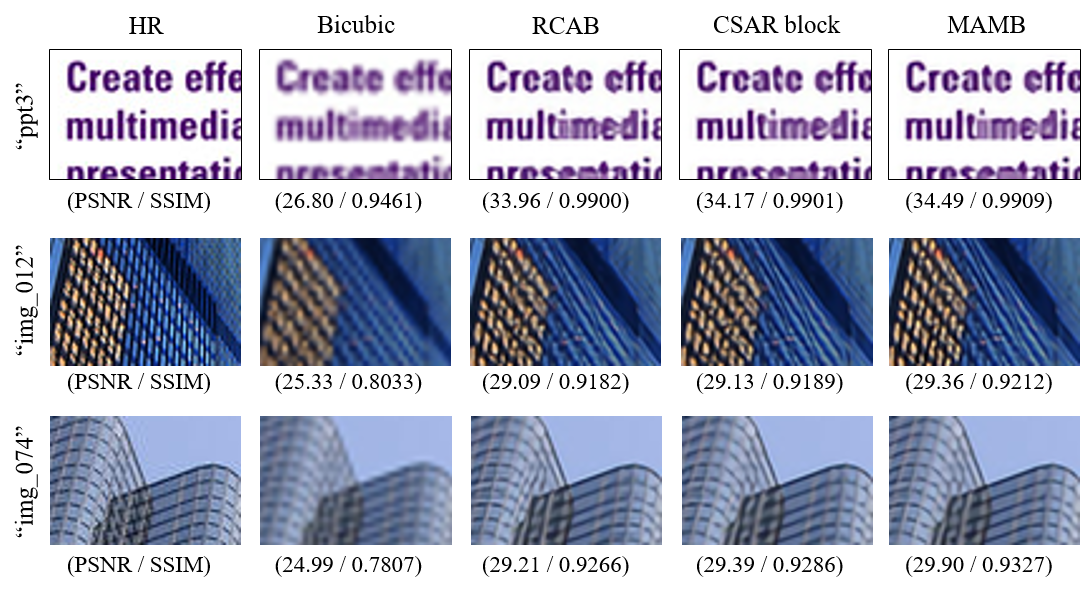}
		\caption{Difficult cases.}
		\label{fig:diff}
	\end{subfigure}
	\caption{Visual comparison of $\times 2$ SR results by our method and the other feature modulation methods on Set14~\cite{zeyde2010single} and Urban100~\cite{huang2015single}.}
	\label{fig:attention_comp2}
\end{figure*}

\subsection{Comparison with Other Feature Modulation Methods}
To demonstrate effectiveness and efficiency of our proposed MAMB, we evaluate our MAMB by comparing with other feature modulation strategies for image SR~\cite{zhang2018rcan,hu2018channel}.
For fairness, we construct networks in the form of implementing the feature modulation methods in each residual block of the baseline network.

\textbf{Quantitative comparison.}
We compare performances across networks of varying sizes ($R$), which are shown in Figure~\ref{fig:attention_comp}.
We can see that our method performs better for all network sizes than the others.
In other words, our network needs only a relatively small number of parameters to achieve the same performance.

\textbf{Qualitative comparison.}
We further provide qualitative comparison of the feature modulation methods for $R$=16 and $C$=64 in Figure~\ref{fig:attention_comp2}.
Our method shows superior performance particularly in difficult cases (Figure~\ref{fig:diff}), while maintaining similar performance with the others in relatively easy cases (Figure~\ref{fig:easy}).
The result shows that our method succeeds in effective feature modulation, thereby improving the ability to adapt to various situations.

\begin{table*}[t]
	\centering
	\scalebox{0.9}{
		\begin{tabular}{clccccc}
			\toprule
			\multirow{2}{*}[-2pt]{Scale} &\multirow{2}{*}[-2pt]{Method} & Set5 & Set14 & BSD100 & Urban100 & Manga109\\
			\cmidrule{3-7}
			& & PSNR / SSIM & PSNR / SSIM & PSNR / SSIM & PSNR / SSIM & PSNR / SSIM \\
			\midrule
			\multirow{11}{*}{$\times$2} & VDSR \cite{kim2016accurate}  & 37.53 / 0.9587 & 33.03 / 0.9124 & 31.90 / 0.8960 & 30.76 / 0.9140 & 37.22 / 0.9750  \\
			& LapSRN \cite{lai2017laplacian}  & 37.52 / 0.9591 & 33.08 / 0.9130 & 31.80 / 0.8950 & 30.41 / 0.9101 & 37.27 / 0.9740  \\
			& DRRN \cite{tai2017image}  & 37.74 / 0.9591 & 33.23 / 0.9136 & 32.05 / 0.8973 & 31.23 / 0.9188 & 37.92 / 0.9760\\
			& MemNet \cite{tai2017memnet}  & 37.78 / \textcolor{black}{0.9597} & 33.23 / 0.9142 & 32.08 / 0.8978 & 31.31 / 0.9195 & 37.72 / 0.9740 \\
			& DSRN \cite{han2018image}  & 37.66 / 0.9590 & 33.15 / 0.9130 & \textcolor{black}{32.10} / 0.8970 & 30.97 / 0.9160 & - / - \\
			& SRMDNF \cite{zhang2018learning} & 37.79 / 0.9601 & 33.32 / 0.9159 & \textcolor{black}{32.05} / 0.8985 & 31.33 / 0.9204 & 38.07 / 0.9761  \\			
			& IDN \cite{hui2018fast} & \textcolor{black}{37.83} / \textcolor{black}{0.9600} & 33.30 / 0.9148 & 32.08 / \textcolor{black}{0.8985} & 31.27 / 0.9196 & - / - \\
			& CARN \cite{ahn2018fast} & 37.76 / 0.9590 & \textcolor{black}{33.52} / \textcolor{black}{0.9166} & 32.09 / 0.8978 & \textcolor{black}{31.92} / \textcolor{black}{0.9256} & - / - \\		
			& MSRN~\cite{li2018multi} & \textcolor{black}{37.90} / \textcolor{black}{0.9597} & 33.62 / 0.9177 & 32.16 / \textcolor{black}{0.8995} & 32.22 / \textcolor{black}{0.9295} & \textcolor{black}{38.40} / \textcolor{black}{0.9761} \\
			& D-DBPN~\cite{haris2018deep} & \textcolor{blue}{38.05} / \textcolor{blue}{0.9599} & \textcolor{blue}{33.79} / \textcolor{blue}{0.9193} & \textcolor{blue}{32.25} / \textcolor{blue}{0.9001} & \textcolor{blue}{32.51} / \textcolor{blue}{0.9317} & \textcolor{blue}{38.81} / \textcolor{blue}{0.9766} \\
			& MAMNet & \textcolor{red}{38.10} / \textcolor{red}{0.9601} & \textcolor{red}{33.90} / \textcolor{red}{0.9199} & \textcolor{red}{32.30} / \textcolor{red}{0.9007} & \textcolor{red}{32.94} / \textcolor{red}{0.9352} & \textcolor{red}{39.15} / \textcolor{red}{0.9772} \\
			\midrule
			
			\multirow{10}{*}{$\times$3} & VDSR \cite{kim2016accurate}  & 33.66 / 0.9213 & 29.77 / 0.8314 & 28.82 / 0.7976 & 27.14 / 0.8279 & 32.01 / 0.9340  \\
			& LapSRN \cite{lai2017laplacian}  & 33.82 / 0.9227 & 29.87 / 0.8320 & 28.82 / 0.7980 & 27.07 /  0.8280 & 32.21 / 0.9350  \\
			& DRRN \cite{tai2017image}  & 34.02 / 0.9244 & 30.08 / 0.8361 & 28.95 / 0.8007 & 27.54 / 0.8378 & 32.72 / 0.9380 \\
			& MemNet \cite{tai2017memnet}  & 34.09 / \textcolor{black}{0.9248} & 30.00 / 0.8350 & 28.96 / 0.8001 & 27.56 / 0.8376 & 32.51 / 0.9369 \\
			& DSRN \cite{han2018image}  & 33.88 / 0.9220 & 30.26 / 0.8370 & \textcolor{black}{28.81} / 0.7970 & 27.16 / 0.8280 & - / - \\
			& SRMDNF \cite{zhang2018learning} & 34.12 / 0.9254 & 30.04 / 0.8382 & \textcolor{black}{28.97} / 0.8025 & 27.57 / 0.8398 & 33.00 / 0.9403  \\
			& IDN \cite{hui2018fast} & \textcolor{black}{34.11} / \textcolor{black}{0.9253} & 29.99 / 0.8354 & 28.95 / \textcolor{black}{0.8013} & 27.42 / 0.8359 & - / - \\
			& CARN \cite{ahn2018fast} & 34.29 / 0.9255 & \textcolor{black}{30.29} / \textcolor{black}{0.8407} & 29.06 / 0.8034 & \textcolor{black}{27.38} / \textcolor{black}{0.8404} & - / - \\		
			& MSRN~\cite{li2018multi} & \textcolor{blue}{34.38} / \textcolor{blue}{0.9265} & \textcolor{blue}{30.37} / \textcolor{blue}{0.8428} & \textcolor{blue}{29.12} / \textcolor{blue}{0.8056} & \textcolor{blue}{28.31} / \textcolor{blue}{0.8553} & \textcolor{blue}{33.59} / \textcolor{blue}{0.9442} \\
			& MAMNet & \textcolor{red}{34.61} / \textcolor{red}{0.9281} & \textcolor{red}{30.54} / \textcolor{red}{0.8459} & \textcolor{red}{29.25} / \textcolor{red}{0.8082} & \textcolor{red}{28.82} / \textcolor{red}{0.8648} & \textcolor{red}{34.14} / \textcolor{red}{0.9472} \\
			\midrule
			
			\multirow{12}{*}{$\times$4} & VDSR \cite{kim2016accurate}  & 31.35 / 0.8838 & 28.01 / 0.7674 & 27.29 / 0.7251 & 25.18 / 0.7524 & 28.83 / 0.8870\\
			& LapSRN \cite{lai2017laplacian} & 31.54 / 0.8850 & 28.19 / 0.7720 & 27.32 / 0.7270 & 25.21 / 0.7560 &29.09 / 0.8900 \\
			& DRRN \cite{tai2017image} & 31.68 / 0.8888 & 28.21 / 0.7720 & 27.38 / 0.7284 & 25.44 / 0.7638 & 29.46 / 0.8960\\
			& MemNet \cite{tai2017memnet}  & 31.74 / 0.8893 & 28.26 / 0.7723 & 27.40 / 0.7281 & 25.50 / 0.7630 & 29.42 / 0.8942 \\
			& SRDenseNet~\cite{tong2017image}  & 32.02 / 0.8934 & 28.50 / 0.7782 & 27.53 / 0.7337 & \textcolor{black}{26.05} / 0.7819 & - / -\\
			& DSRN \cite{han2018image}  & 31.40 / 0.8830 & 28.07 / 0.7700 & 27.25 / 0.7240 & 25.08 / 0.7170 & - / -\\
			& SRMDNF \cite{zhang2018learning}  & 31.96 / 0.8925 & 28.35 / 0.7787 & 27.49 / 0.7337 & 25.68 / 0.7731 & 30.09 / 0.9024 \\			
			& IDN \cite{hui2018fast}  & 31.82 / 0.8903 & 28.25 / 0.7730 & 27.41 / 0.7297 & 25.41 / 0.7632 & - / -\\
			& CARN \cite{ahn2018fast}  & \textcolor{black}{32.13} / \textcolor{black}{0.8937} & \textcolor{black}{28.60} / \textcolor{black}{0.7806} & \textcolor{black}{27.58} / \textcolor{black}{0.7349} & \textcolor{black}{26.07} / \textcolor{black}{0.7837} & - / -\\
			& MSRN~\cite{li2018multi}  & \textcolor{black}{32.21} / \textcolor{black}{0.8949} & 28.61 / 0.7827 & 27.60 / \textcolor{black}{0.7372} & 26.20 / \textcolor{black}{0.7903} & \textcolor{black}{30.53} / \textcolor{black}{0.9090} \\
			& D-DBPN~\cite{haris2018deep}  & \textcolor{blue}{32.40} / \textcolor{blue}{0.8966} & \textcolor{blue}{28.75} / \textcolor{red}{0.7854} & \textcolor{blue}{27.67} / \textcolor{blue}{0.7385} & \textcolor{blue}{26.38} / \textcolor{blue}{0.7938} & \textcolor{blue}{30.89} / \textcolor{blue}{0.9127} \\
			& MAMNet & \textcolor{red}{32.42} / \textcolor{red}{0.8972} & \textcolor{red}{28.77} / \textcolor{red}{0.7854} & \textcolor{red}{27.70} / \textcolor{red}{0.7406} & \textcolor{red}{26.59} / \textcolor{red}{0.8013} & \textcolor{red}{30.94} / \textcolor{red}{0.9142} \\
			\bottomrule
		\end{tabular}
	}
	\caption{Quantitative evaluation results of SR models for scaling factors of 2, 3 and 4. Red and blue colors indicate the best and second best performance, respectively.}
	\label{tab:MSRN}
\end{table*}

\begin{figure*}[t]
	\centering
	\includegraphics[width=0.9\textwidth]{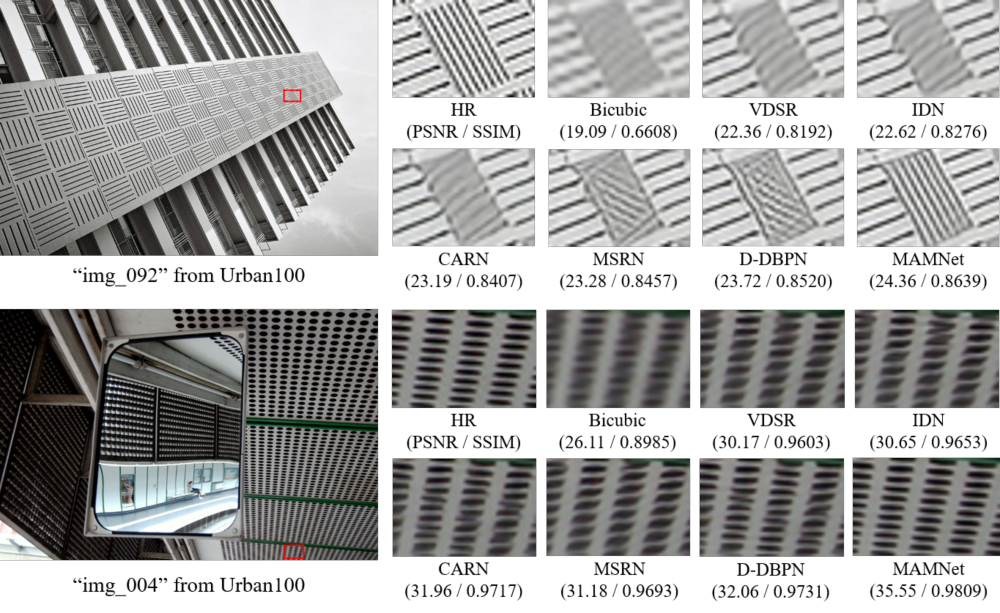}
	\caption{Visual comparison of $\times 2$ SR results on Urban100~\cite{huang2015single}.}
	\label{fig:result1}
\end{figure*}

\begin{figure*}[t]
	\centering
	\includegraphics[width=0.9\textwidth]{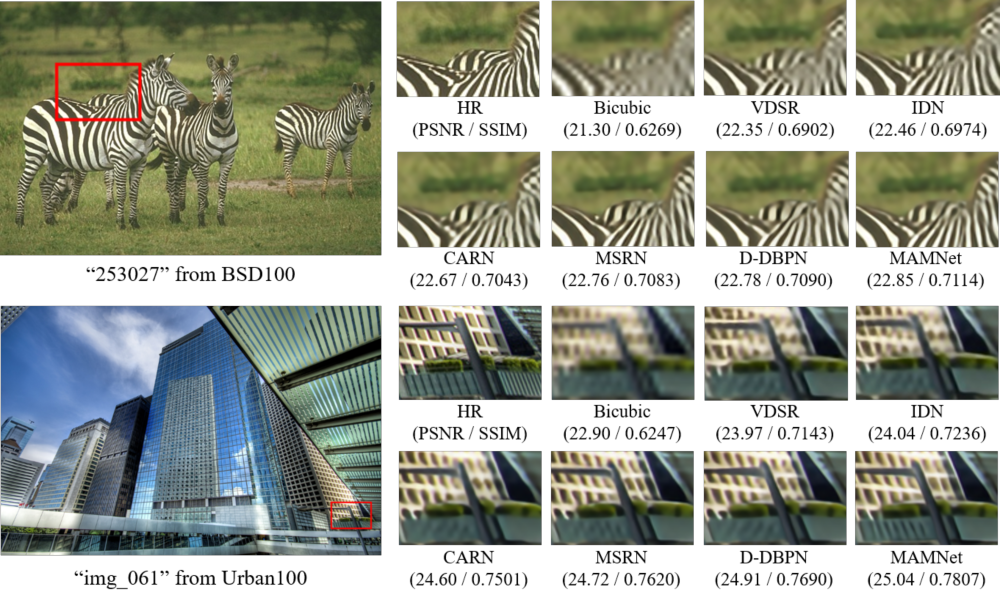}
	\caption{Visual comparison of $\times 4$ SR results on BSD100~\cite{martin2001database} and Urban100~\cite{huang2015single}.}
	\label{fig:result2}
\end{figure*}

\begin{figure*}[t]
	\centering
	\includegraphics[width=0.9\textwidth]{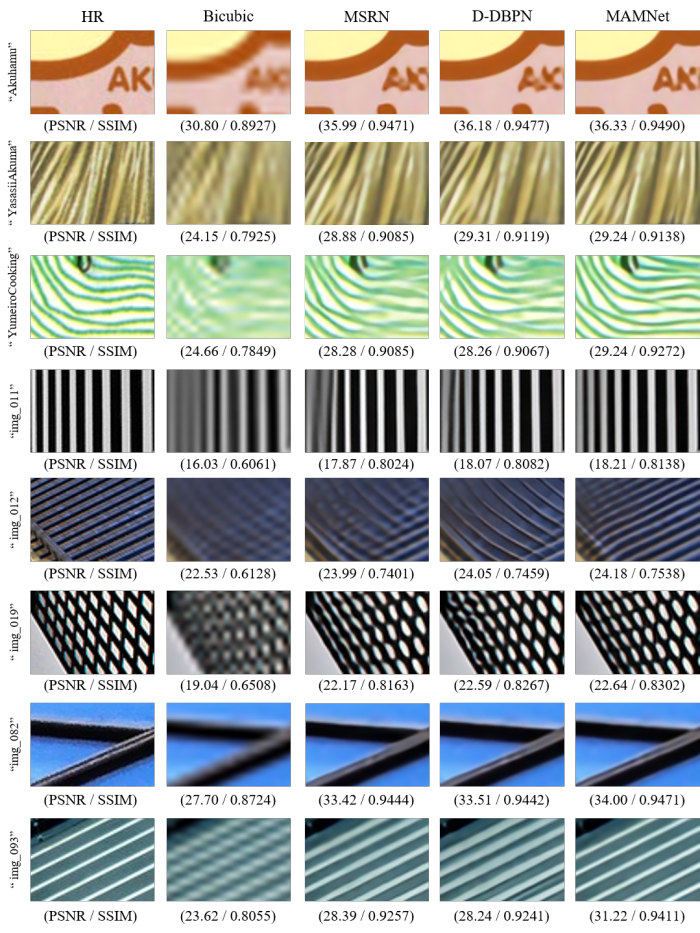}
	\caption{Visual comparison of $\times 4$ SR results on the challenging images of Urban100~\cite{huang2015single} and Manga109~\cite{matsui2017sketch}.}
	\label{fig:result3}
\end{figure*}

\begin{figure}[t]
	\centering
	\begin{subfigure}[b]{0.405\textwidth}
		\includegraphics[width=\textwidth]{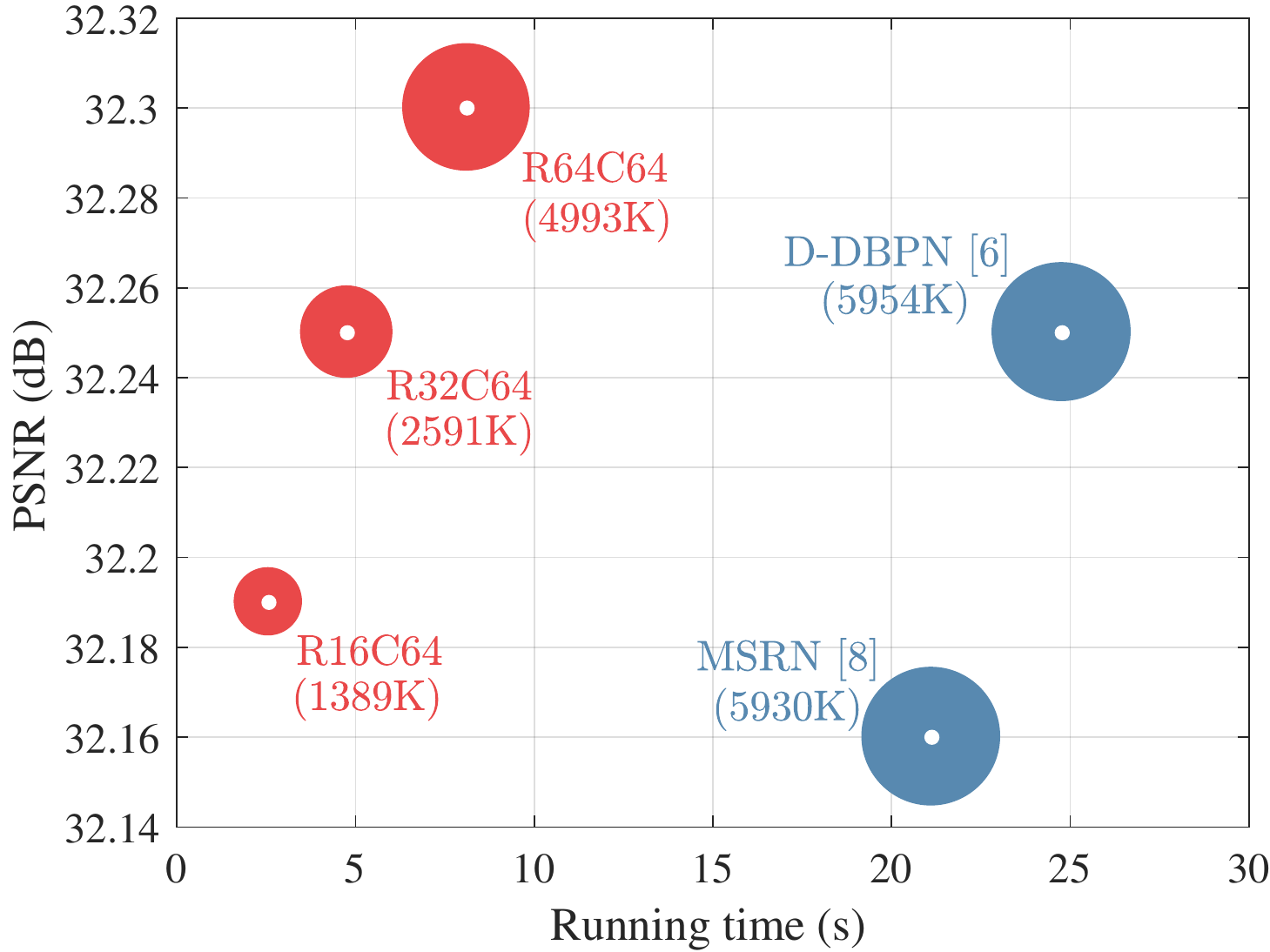}
		\caption{BSD100}
	\end{subfigure}
	\begin{subfigure}[b]{0.4\textwidth}
		\includegraphics[width=\textwidth]{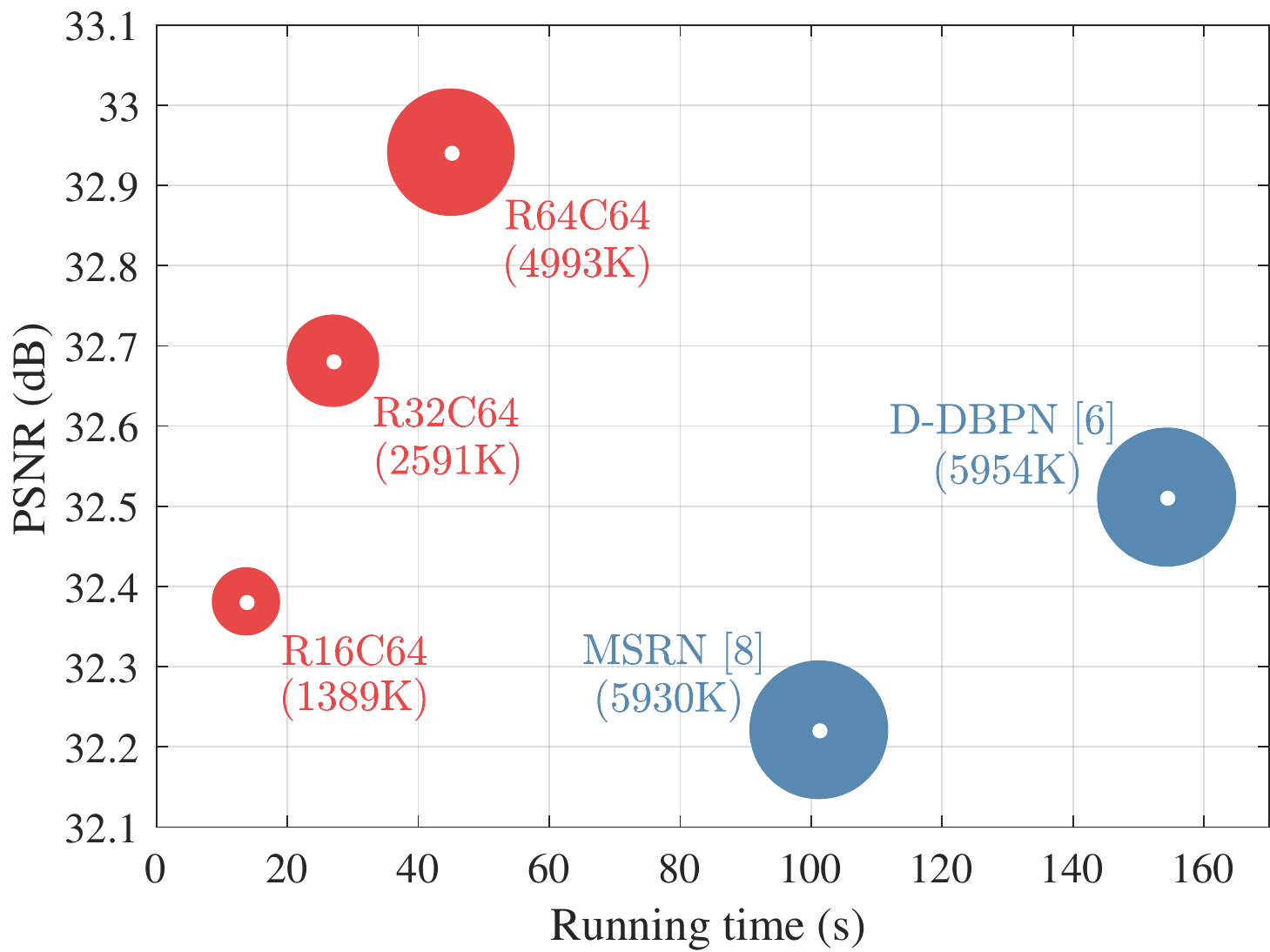}
		\caption{Urban100}
	\end{subfigure} 
	\caption{PSNR (dB) vs. running time (s) for $\times$2 SR. The PSNR values and running times are average values for each dataset. Two existing methods are shown with blue color, and our models with varying the number of parameters ($R$) are shown with red color. The area of each circle is proportional to the number of parameters in each model (also shown as numbers in parentheses). 
	}
	\label{fig:result2}	
\end{figure}

\subsection{Comparison with State-of-the-art Methods}
\textbf{Quantitative and qualitative comparisons.}
We finally evaluate our proposed MAMNet by comparing with 11 state-of-the art SR methods: VDSR~\cite{kim2016accurate}, LapSRN~\cite{lai2017laplacian}, DRRN~\cite{tai2017image}, MemNet~\cite{tai2017memnet}, SRDenseNet~\cite{tong2017image}, DSRN~\cite{han2018image}, SRMDNF~\cite{zhang2018learning}, IDN~\cite{hui2018fast}, CARN~\cite{ahn2018fast}, MSRN~\cite{li2018multi}, and D-DBPN~\cite{haris2018deep}.
Here, EDSR~\cite{li2018multi}, RDN~\cite{zhang2018residual}, and RCAN~\cite{zhang2018rcan} are excluded from the comparison because they have extremely larger numbers of parameters.
We select MAMNet with $R=64$ and $C=64$ (MAMNet\_$R64C64$) as our final model.
The $\times$2, $\times$3 and $\times$4 SR quantitative results are summarized in Table~\ref{tab:MSRN}.
MAMNet shows the highest PSNR and SSIM values on all datasets for all scaling factors, and the performance gap with the other methods is particularly prominent in Urban100 and Manga109.
These results demonstrate the effectiveness of our proposed MAMNet.

We also provide the visual results of $\times$2 super-resolved images in Figure~\ref{fig:result1}, where only our model successfully restores complex patterns.
For ``img\_092'' from Urban100, our proposed method takes advantage of the peripheral information more actively, i.e., the information about the repeated pattern.
Similarly, in ``img\_004'', it can be seen that the learning of the repeated pattern is not performed well by merely using the local information from the LR image. 
On the other hand, our model recovers the pattern correctly.
Furthermore, we show the $\times$4 super-resolved images from BSD100 and Urban100 in Figure~\ref{fig:result2}.
For the ``253027'' image, we can see that our network expresses the complicated stripes more finely.
For ``img\_061'', the outputs of the other models look blurry or have patterns in wrong directions, while only our MAMNet restores the correct pattern.

Visual results of challenging images for $\times$4 SR are shown in Figure~\ref{fig:result3}.
Here, we focus on comparison with the top two models, MSRN~\cite{li2018multi} and D-DBPN~\cite{haris2018deep}.
For image ``Akuhamu'', MAMNet restores better the densely written letters, ``A'' and ``K''.
For images ``YasasiiAkuma'' and ``YumeiroCooking'', MAMNet reconstructs complicated straight lines and curves more clearly. 
For images ``img\_011'', ``img\_012'', ``img\_019'', and ``img\_093'', while the others fails to restore the patterns in terms of thickness, direction, and spacing, MAMNet successfully generates the repeated patterns. 
For image ``img\_082'', MAMNet restores black sharp lines relatively well.
These results demonstrate the strength of our proposed method in various difficult SR tasks.



\textbf{Model efficiency.}
MAMNet enables highly effective and efficient SR through multi-path adaptive modulation.
It is powerful, but relatively lightweight and fast compared to the other state-of-the art models.
We show the efficiency of our MAMNet in Figure~\ref{fig:result2}.
MAMNet\_$R64C64$ show the best performance on both datasets with smaller numbers of parameters and shorter running time than D-DBPN and MSRN.
MAMNet\_$R32C64$ has similar or better performance on both datasets with only 43.69$\%$ and 43.52$\%$ of the number of parameters of MSRN and D-DBPN, respectively.
In addition, its running time is only 22.59$\%$ (BSD100) and 26.76$\%$ (Urban100) of that of MSRN, and 19.26$\%$ (BSD100) and 17.56$\%$ (Urban100) of that of D-DBPN. 
For both datasets, MAMNet\_$R16C64$ outperforms MSRN with only 23.42$\%$ parameters of MSRN.
It takes only 12.20$\%$ (BSD100) and 13.66$\%$ (Urban100) of the running time of MSRN.

\section{Conclusion}
\label{section:conclusion}
In this paper, we proposed a novel multi-path adaptive modulation network (MAMNet) for image SR. 
We proposed three feature modulation methods by exploiting the convolutional feature responses effectively.
We demonstrated that the proposed MAMB is more effective and parameter-efficient than existing feature modulation methods for SR.
The experimental results also demonstrated that the proposed MAMNet can achieve improved SR performance compared to the state-of-the-art methods with a relatively small number of parameters.


\section*{Acknowledgements}
This research was supported by the MSIT (Ministry of Science and ICT), Korea, under the ``ICT Consilience Creative Program'' (IITP-2018-2017-0-01015) supervised by the IITP (Institute for Information \& communications Technology Promotion). In addition, this work was also supported by the IITP grant funded by the Korea government (MSIT) (R7124-16-0004, Development of Intelligent Interaction Technology Based on Context Awareness and Human Intention Understanding).

\section*{References}
\bibliography{neurocomputing}
\end{document}